\documentclass[a4paper]{article}

\usepackage[english]{babel}
\usepackage[utf8x]{inputenc}
\usepackage[T1]{fontenc}

\usepackage[a4paper,top=3cm,bottom=2cm,left=3cm,right=3cm,marginparwidth=1.75cm]{geometry}

\usepackage{amsmath}
\usepackage{graphicx}
\usepackage[colorinlistoftodos]{todonotes}
\usepackage[colorlinks=true, allcolors=blue]{hyperref}
\usepackage{enumitem}
\usepackage[toc,page]{appendix}
\usepackage{authblk}

\def\boxitem#1{\setbox0=\vbox{#1}{\centering\makebox[0pt]{%
  \fboxrule=1pt\color{black}\fbox{\hspace{\leftmargini}\color{black}\box0}}\par}}

\title{\textbf{Model Trees for Identifying Exceptional Players in the NHL Draft}}
\author[1]{Oliver Schulte}
\author[2]{Yejia Liu}
\author[3]{Chao Li}
\affil[1,2,3]{Department of Computing Science, Simon Fraser University}
\affil[ ]{\textit {\{oschulte, yejial, chao\_li\_2\}@sfu.ca}}

\begin{document}
\maketitle

\begin{abstract}
Drafting strong players is crucial for a team’s success. We describe a new data-driven interpretable approach for assessing draft prospects in the National Hockey League. Successful previous approaches have 1) built a predictive model based on player features (e.g. Schuckers 2017 \cite{Schuckers2016}, or 2) derived performance predictions from the observed performance of comparable players in a cohort (Weissbock 2015 \cite{PCS}). This paper develops model tree learning, which incorporates strengths of both model-based and cohort-based approaches. A model tree partitions the feature space according to the values of discrete features, or learned thresholds for continuous features. Each leaf node in the tree defines a group of players, easily described to hockey experts, with its own group regression model. Compared to a single model, the model tree forms an ensemble that increases predictive power. Compared to cohort-based approaches, the groups of comparables are discovered from the data, without requiring a similarity metric. The performance predictions of the model tree are competitive with the state-of-the-art methods, which validates our model empirically. We show in case studies that the model tree player ranking can be used to highlight strong and weak points of players.

\end{abstract}

\section{Introduction}
Player ranking is one of the most studied subjects in sports analytics \cite{Swartz}. In this paper we consider predicting success in the National Hockey League(NHL) from junior league data, with the goal of supporting draft decisions. The publicly available junior league data aggregate a season’s performance into a single set of numbers for each player.  Our method can be applied to any data of this type, for example also to basketball NBA draft data(\url{www.basketball-reference.com/draft/}). Since our goal is to support draft decisions by teams, we ensure that the results of our data analysis method can be easily explained to and interpreted by sports experts. 

Previous approaches for analyzing hockey draft data take a regression approach or a similarity-based approach. Regression approaches build a predictive model that takes as input a set of player features, such as demographics (age, height, weight) and junior league performance metrics (goals scored, plus-minus), and output a predicted success metric (e.g. number of games played in the professional league). The current state-of-the-art is a generalized additive model \cite{Schuckers2016}. Cohort-based approaches divide players into groups of comparables and predict future success based on a player’s cohort. For example, the PCS model \cite{PCS} clusters players according to age, height, and scoring rates. One advantage of the cohort model is that predictions can be explained by reference to similar known players, which many domain experts find intuitive. For this reason, several commercial sports analytics systems, such as Sony’s Hawk-Eye system, identify groups of comparables for each player. Our aim in this paper is to describe a new model for draft data that achieves the best of both approaches, regression-based and similarity-based. 

Our method uses a model tree \cite{Friedman00, GUIDE}. Each node in the tree defines a new yes/no question, until a leaf is reached. Depending on the answers to the questions, each player is assigned a group corresponding to a leaf. The tree builds a different regression model for each leaf node. Figure 1 shows an example model tree. A model tree offers several advantages.

\begin{figure}[!h]
    \begin{center}
        \includegraphics[width=1.0\textwidth]{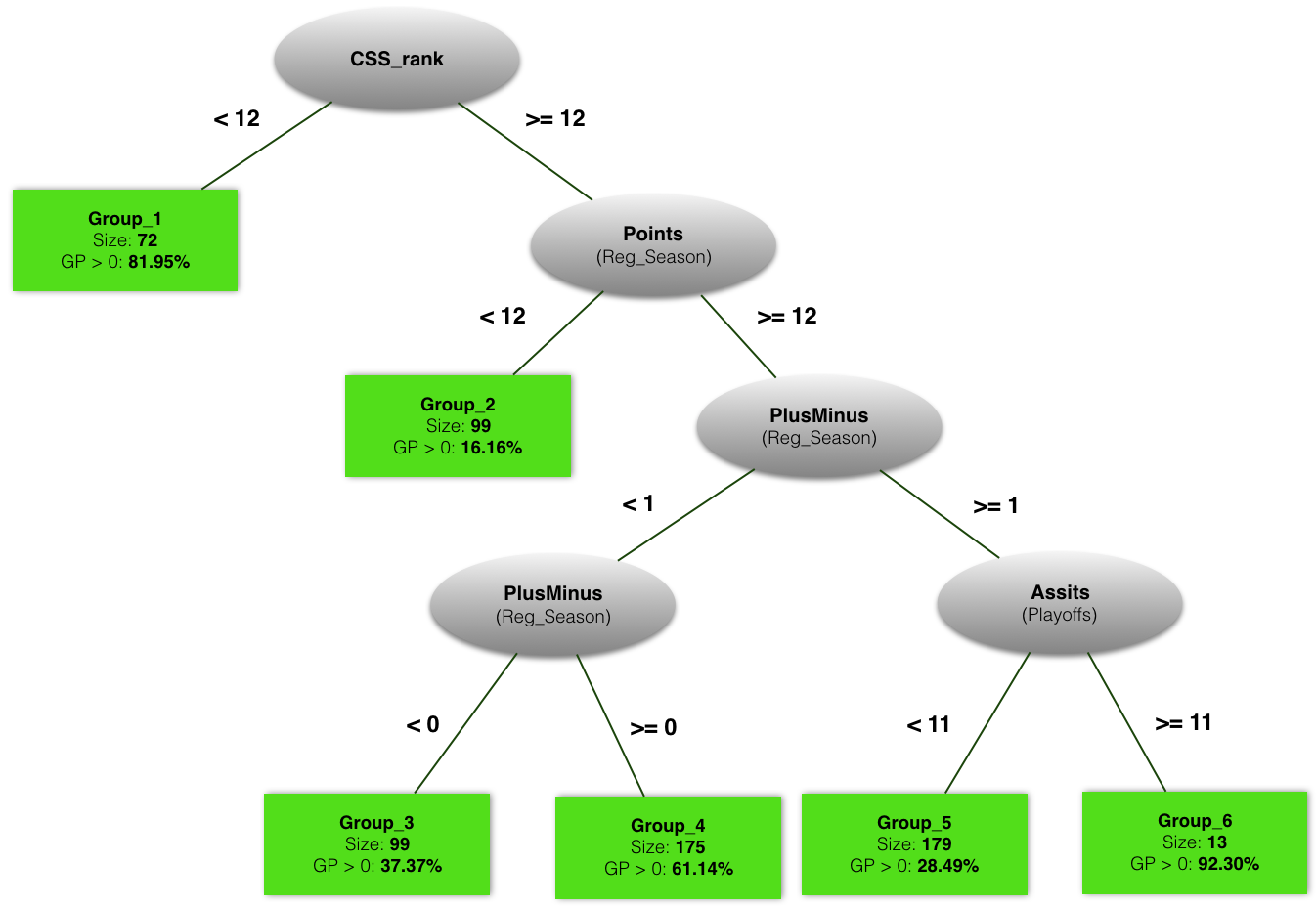}
    \caption{Logistic Regression Model Trees for the ${2004, 2005, 2006}$ cohort in NHL. The tree was built using the LogitBoost algorithm implemented in the LMT package of the Weka Program \cite{Weka1, Hall2}.}
    \end{center}
    \end{figure}

\begin{itemize}
\item {Compared to a single regression model, the tree defines an ensemble of regression models, based on non-linear thresholds. This increases the expressive power and predictive accuracy of the model. The tree can represent complex interactions between player features and player groups. For example, if the data indicate that players from different junior leagues are sufficiently different to warrant building distinct models, the tree can introduce a split to distinguish different leagues.}

\item {Compare to a similarity-based model, tree construction learns groups of players from the data, without requiring the analyst to specify a similarity metric. Because tree learning selects splits that increase predictive accuracy, the learned distinctions between the groups are guaranteed to be predictively relevant to future NHL success. Also, the tree creates a model, not a single prediction, for each group, which allows it to differentiate players from the same group.}
\end{itemize}

A natural approach would be to build a linear regression tree to predict NHL success, which could be measured by the number of games a draft pick plays in the NHL. However, only about half the draft picks ever play a game in the NHL \cite{Tingling}. As observed by \cite{Schuckers2016}, this creates a zero-inflation problem that limits the predictive power of linear regression. We propose a novel solution to the zero-inflation problem, which applies logistic regression to predict whether a player will play at least one game in the NHL. We learn a logistic regression model tree, and rank players by the probability that the logistic regression model tree assigns to them playing at least one game. Intuitively, if we can be confident that a player will play at least one NHL game, we can also expect the player to play many NHL games. Empirically, we found that on the NHL draft data, the logistic regression tree produces a much more accurate player ranking than the linear regression tree. 

Following \cite{Schuckers2016}, we evaluate the logistic regression ranking by comparing it to ranking players by their future success, measured as the number of NHL games they play after 7 years. The correlation of the logistic regression ranking with future success is competitive with that achieved by the generalized additive model of \cite{Schuckers2016}. We show in case studies that the logistic model tree adds information to the NHL’s Central Scouting Service Rank (CSS). For example, Stanley Cup winner \textit{Kyle Cumiskey} was not ranked by the CSS in his draft year, but was ranked as the third draft prospect in his group by the model tree, just behind \textit{Brad Marchand} and \textit{Mathieu Carle}. Our case studies also show that the feature weights learned from the data can be used to explain the ranking in terms of which player features contribute the most to an above-average ranking. In this way the model tree can be used to highlight exceptional features of a player for scouts and teams to take into account in their evaluation. 

\textit{Paper Outline.} After we review related work, we show and discuss the model tree learned from the 2004-2006 draft data. The rank correlations are reported to evaluate predictive accuracy. We discuss in detail how the ensemble of group models represents a rich set of interactions between player features, player categories, and NHL success. Case studies give examples of strong players in different groups and show how the model can used to highlight exceptional player features.

\section{Related Work}
Different approaches to player ranking are appropriate for different data types. For example, with dynamic play-by-play data, Markov models have been used to rank players \cite{Cervone2014,Thomas2013,Oliver2017,Kaplan2014}. For data that record the presence of players when a goal is scored, regression models have also been applied to extend the classic plus-minus metric \cite{Macdonald2011,Gramacy2013}. In this paper, we utilize player statistics that aggregate a season's performance into a single set of numbers. While this data is much less informative than play-by-play data, it is easier to obtain, interpret, and process. 

\textit{Regression Approaches.} To our knowledge, this is the first application of model trees to hockey draft prediction, and the first model for predicting whether a draftee plays any games at all. The closest predecessor to our work is due to Schuckers \cite{Schuckers2016}, who uses a single generalized additive model to predict future NHL game counts from junior league data. 

\textit{Similarity-Based Approaches} assume a similarity metric and group similar players to predict performance. A sophisticated example from baseball is the nearest neighbour analysis in the PECOTA system \cite{PECOTA}. For ice hockey, the Prospect Cohort Success (PCS) model \cite{PCS}, cohorts of draftees are defined based on age, height, and scoring rates. Model tree learning provides an automatic method for identifying cohorts with predictive validity.  We refer to cohorts as groups to avoid confusion with the PCS concept. Because tree learning is computationally efficient, our model tree is able to take into account a larger set of features than age, height, and scoring rates. Also, it provides a separate predictive model for each group that assigns group-specific weights to different features. In contrast, PCS makes the same prediction for all players in the same cohort. So far, PCS has been applied to predict whether a player will score more than 200 games career total. Tree learning can easily be modified to make predictions for any game count threshold.  

\section{Dataset}

Our data was obtained from public-domain on-line sources, including \url{nhl.com}, \url{eliteprospects.com}, and \url{draftanalyst.com}. We are also indebted to David Wilson for sharing his NHL performance dataset \cite{Wilson2016}. The full dataset is posted on the Github(\url{https://github.com/liuyejia/Model_Trees_Full_Dataset}). We consider players drafted into the NHL between 1998 to 2008 (excluding goalies). Following \cite{Schuckers2016}, we took as our dependent variable \textbf{the total number of games $g_i$ played} by a player $i$ after 7 years under an NHL contract. The first seven seasons are chosen because NHL teams have at least seven-year rights to players after they are drafted \cite{Schucker2013}. Our dataset includes also the total time on ice after $7$ years. The results for time on ice were very similar to number of games, so we discuss only the results for number of games. The independent variables include demographic factors (e.g. age), performance metrics for the year in which a player was drafted (e.g., goals scored), and the rank assigned to a player by the NHL Central Scouting Service (CSS). If a player was not ranked by the CSS, we assigned (1+ the maximum rank for his draft year) to his CSS rank value. Another preprocessing step was to pool all European countries into a single category. If a player played for more than one team in his draft year (e.g., a league team and a national team), we added up this counts from different teams. Table 1 lists all data columns and their meaning. Figure 1 shows an excerpt from the dataset. 

\begin{table}[!h]
    \begin{center}
    \begin{tabular}{ | l | p{10cm} |}
    \hline
    Variable Name & Description \\ \hline
    id & nhl.com id for NHL players, otherwise Eliteprospects.com id \\ \hline
    
    DraftAge & Age in Draft Year \\ \hline
    
    Country & Nationality. Canada -> 'CAN', USA -> 'USA', countries in Europe -> 'EURO' \\ \hline
    
    Position & Position in Draft Year. Left Wing -> 'L', Right Wing -> 'R', Center -> 'C', Defencemen -> 'D' \\ \hline
    
    Overall & Overall pick in NHL Entry Draft \\ \hline
    
    CSS\_rank & Central scouting service ranking in Draft Year \\ \hline
    rs\_GP   & Games played in regular seasons in Draft Year \\ \hline
    rs\_G & Goals in regular seasons in Draft Year \\ \hline
    rs\_A & Assists in regular seasons in Draft Year \\ \hline
    rs\_P & Points in regular seasons in Draft Year \\ \hline
    rs\_PIM & Penalty Minutes in regular seasons in Draft Year \\ \hline
    rs\_PlusMinus & Goal Differential in regular seasons in Draft Year\\ \hline
    po\_GP & Games played in playoffs in Draft Year \\ \hline
    po\_G & Goals in playoffs in Draft Year \\ \hline
    po\_A & Assists in playoffs in Draft Year \\ \hline
    po\_P & Points in playoffs in Draft Year \\ \hline
    po\_PIM & Penalty Minutes in playoffs in Draft Year \\ \hline
    po\_PlusMinus & Goal differential in playoffs in Draft Year \\ \hline
    sum\_7yr\_GP & Total NHL games played in player's first 7 years of NHL career \\ \hline
    sum\_7yr\_TOI & Total NHL Time on Ice in player's first 7 years of NHL career \\ \hline
    GP\_7yr\_greater\_than\_0 & Played a game or not in player's first 7 years of NHL career\\ \hline

    \end{tabular}
    \caption{Player Attributes listed in dataset \textit{(excluding weight and height)}.}
    \end{center}
    \end{table}
    
    \begin{figure}[!h]
    \begin{center}
        \includegraphics[width=0.9\textwidth]{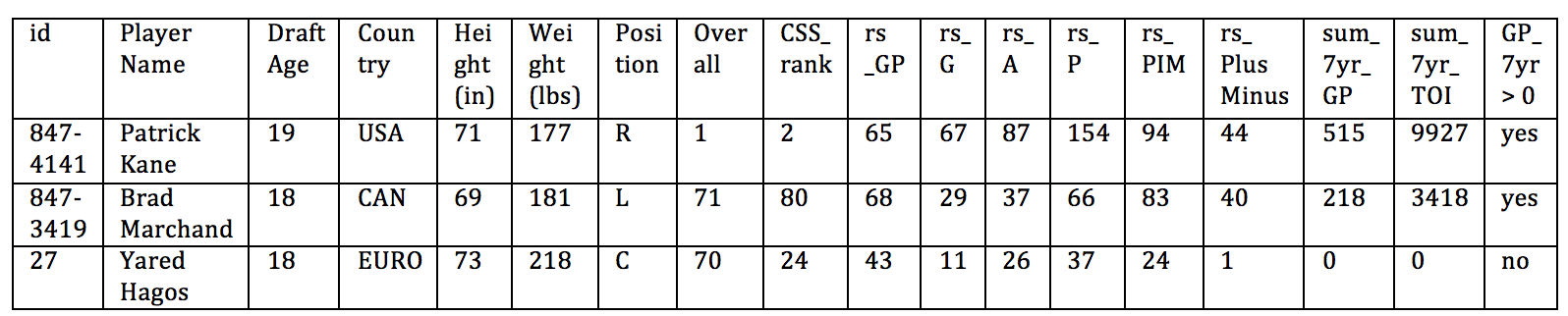}
    \caption{Sample Player Data for their draft year. rs = regular season. We use the same statistics for the playoffs \textit{(not shown)}.}
\end{center}
\end{figure}

\section{Model Tree Construction}

Model trees are a flexible formalism that can be built for any regression model. An obvious candidate for a regression model would be linear regression; alternatives include a generalized additive model \cite{Schuckers2016}, and a Poisson regression model specially built for predicting counts \cite{Ryder}. We introduce a different approach: a logistic regression model to predict whether a player will play any games at all in the NHL ($g_i>0$). The motivation is that many players in the draft never play any NHL games at all (up to 50\% depending on the draft year) \cite{Tingling}. This poses an extreme zero-inflation problem for any regression model that aims to predict directly the number of games played. In contrast, for the classification problem of predicting whether a player will play any NHL games, zero-inflation means that the data set is balanced between the classes. This classification problem is interesting in itself; for instance, a player agent would be keen to know what chances their client has to participate in the NHL. The logistic regression probabilities $p_i=P(g_i>0)$ can be used not only to predict whether a player will play any NHL games, but also to rank players such that the ranking correlates well with the actual number of games played. Our method is therefore summarized as follows.

\begin{enumerate}
    \boxitem{
    \item[1.]
    Build a tree whose leaves contain a logistic regression model.
    \item[2.]
    The tree assigns each player $i$ to a unique leaf node $l_i$, with a logistic regression model $m(l_i)$.
    \item[3.]
    Use $m(l_i)$ to compute a probability $p_i= P(g_i>0)$.
    }
    \end{enumerate}
    
Figure 1 shows the logistic regression model tree learned for our second cohort by the LogiBoost algorithm. It places CSS rank at the root as the most important attribute. Players ranked better than $12$ form an elite group, of whom almost $82\%$ play at least one NHL games. For players at rank $12$ or below, the tree considers next their regular season points total. Players with rank and total points below $12$ form an unpromising group: only $16\%$ of them play an NHL game. Players with rank below $12$ but whose points total is $12$ or higher, are divided by the tree into three groups according to whether their regular season plus-minus score is positive, negative, or $0$. (A three-way split is represented by two binary splits). If the plus-minus score is negative, the prospects of playing an NHL game are fairly low at about $37\%$. For a neutral plus-minus score, this increases to $61\%$. For players with a positive plus-minus score, the tree uses the number of playoff assists as the next most important attribute. Players with a positive plus-minus score and more than $10$ playoff assists form a small but strong group that is $92\%$ likely to play at least one NHL game.

\section{Results: Predictive Modelling}

Following \cite{Schuckers2016}, we evaluated the predictive accuracy of the LMT model using the Spearman Rank Correlation(SRC) between two player rankings: $i)$ the performance ranking based on the actual number of NHL games that a player played, and $ii)$ the ranking of players based on the probability $pi$ of playing at least one game(Tree Model SRC). We also compared it with $iii)$ the ranking of players based on the order in which they were drafted (Draft Order SRC). The draft order can be viewed as the ranking that reflects the judgment of NHL teams. We provide the formula for the Spearman correlation in the Appendix. Table 2 shows the Spearman correlation for different rankings.

    \begin{table}[!h]
        \centering
        \begin{tabular}{|l|c|c|c|r|}
        \hline
        \begin{tabular}{@{}c@{}} Training Data \\  NHL Draft Years \end{tabular} & \begin{tabular}{@{}c@{}} Out of Sample \\  Draft Years\end{tabular}   & \begin{tabular}{@{}c@{}} Draft Order \\  SRC\end{tabular} & \begin{tabular}{@{}c@{}} LMT \\  Classification Accuracy\end{tabular} & \begin{tabular}{@{}c@{}} LMT \\  SRC\end{tabular} \\ \hline 
        1998, 1999, 2000 & 2001 & 0.43 & 82.27\% & 0.83 \\ \hline
        1998, 1999, 2000 & 2002 & 0.30 & 85.79\% & 0.85 \\ \hline
        2004, 2005, 2006 & 2007 & 0.46 & 81.23\% & 0.84 \\ \hline
        2004, 2005, 2006 & 2008 & 0.51 & 63.56\% & 0.71 \\ \hline

        \end{tabular}
        \caption{Predictive Performance (our Logitic Model Trees, over all draft ranking) using Spearman Rank Correlation. Bold indicates the best values.}
       
    \end{table}

\textit{Other Approaches.} We also tried designs based on a linear regression model tree, using the M5P algorithm implemented in the Weka program. The result is a decision stump that splits on CSS rank only, which had substantially worse predictive performance(i.e., Spearman correlation of only $0.4$ for the $2004-2006$ cohort). For the generalized additive model (gam), the reported correlations were $2001: 0.53, 2002: 0.54, 2007: 0.69, 2008: 0.71$ \cite{Schuckers2016}. Our correlation is not directly comparable to the gam model because of differences in data preparation: the gam model was applied only to drafted players who played at least one NHL game, and the CSS rank was replaced by the Cescin conversion factors: for North American players, multiply CSS rank by $1.35$, and for European players, by $6.27$ \cite{Fyffe}. The Cescin conversion factors represent an interaction between the player's country and the player's CSS rank. A model tree offers another approach to representing such interactions: by splitting on the player location node, the tree can build a different model for each location. Whether the data warrant building different models for different locations is a data-driven decision made by the tree building algorithm. The same point applies to other sources of variability, for example the draft year or the junior league. Including the junior league as a feature has the potential to lead to insights about the differences between leagues, but would make the tree more difficult to interpret; we leave this topic for future work. In the next section we examine the interaction effects captured by the model tree in the different models learned in each leaf node. 

\section{Results: Learned Groups and Logistic Regression Models}

We examine the learned group regression models, first in terms of the dependent success variable, then in terms of the player features.

\subsection{Groups and the Dependent Variable}

Figure 3 shows boxplots for the distribution of our dependent variable $g_i$. The strongest groups are, in order, 1, 6, and 4. The other groups show weaker performance on the whole, although in each group some players reach high numbers of games. Most players in Group 2\&3\&4\&5 have GP equals to zero while Group 1\&6 represent the strongest cohort in our prediction, where over $80\%$ players played at least 1 game in NHL. The tree identifies that among the players who do not have a very high CSS rank (worse than $12$), the combination of regular season $Points >= 12$, $PlusMinus > 0$, and $play-off Assists > 10$ is a strong indicator of playing a substantive number of NHL games (median $g_i = 128$).

\begin{figure}[!h]
    \begin{center}
        \includegraphics[width=0.9\textwidth]{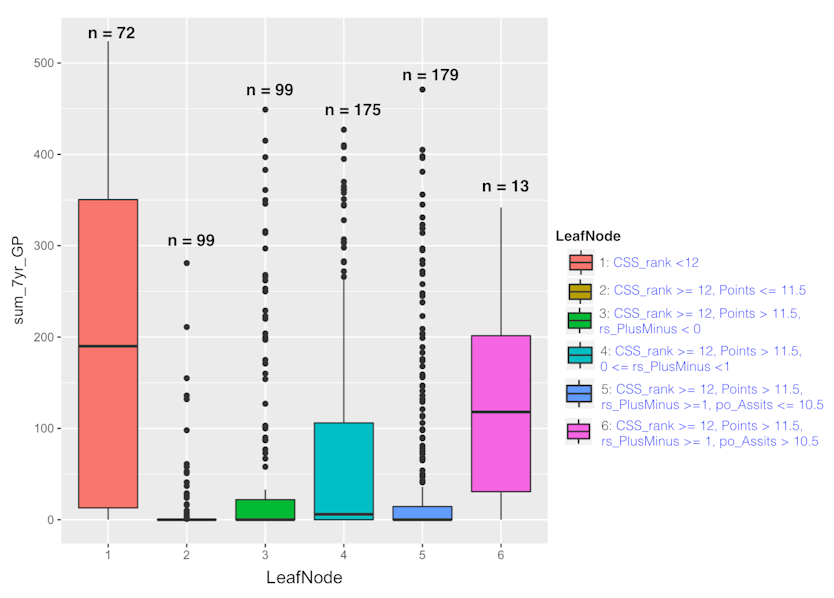}
    \caption{Boxplots for the dependent variable $g_i$ , the total number of NHL games played after $7$ years under an NHL contract. Each boxplot shows the distribution for one of the groups learned by the logistic regression model tree. The group size is denoted $n$.}
    \end{center}
    \end{figure}
    
\subsection{Groups and the Independent Variables}

Figure 5.2 shows the average statistics by group and for all players. The CSS rank for Group 1 is by far the highest. The data validate the high ranking in that $82\%$ players in this group went on to play an NHL game. Group 6 in fact attains an even higher proportion of $92\%$. The average statistics of this group are even more impressive than those of group 1 (e.g., $67$ regular season points in group $6$ vs. $47$ for group 1). But the average CSS rank is the lowest of all groups. So this group may represent a small group of players ($n = 13$) overlooked by the scouts but identified by the tree. Other than Group 6, the group with the lowest CSS rank on average is Group 2. The data validate the low ranking in that only $16\%$ of players in this group went on to play an NHL game. The group averages are also low (e.g., $6$ regular season points is much lower than other groups). 

\begin{figure}[!h]
    \begin{center}
        \includegraphics[width=0.9\textwidth]{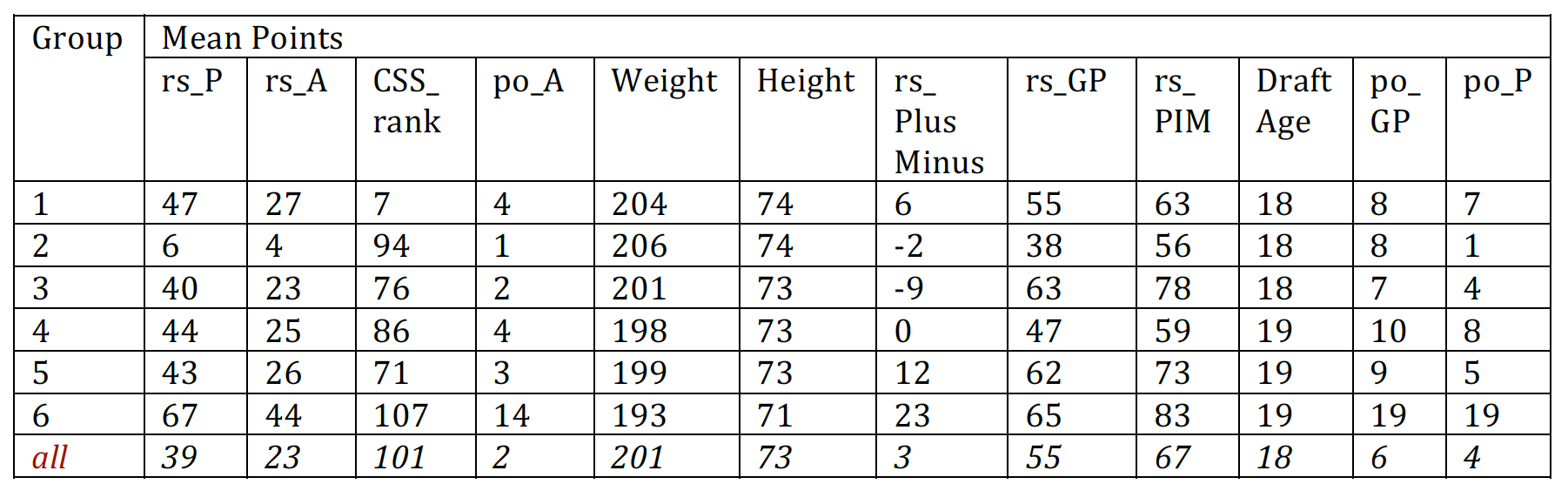}
    \caption{Statistics for the average players in each group and all players.}
    \end{center}
 \end{figure}
 
\section{Group Models and Variable Interactions}

Figure 5 illustrates logistic regression weights by group. A positive weight implies that an increase in the covariate value predicts a large increase in the probability of playing more than one game, compared to the probability of playing zero games. Conversely, a negative weight implies that an increase in the covariate value decreases the predicted probability of playing more than one game. Bold numbers show the groups for which an attribute is most relevant. The table exhibits many interesting interactions among the independent variables; we discuss only a few. Notice that if the tree splits on an attribute, the attribute is assigned a high-magnitude regression weight by the logistic regression model for the relevant group. Therefore our discussion focuses on the tree attributes.
    
    \begin{figure}[!h]
    \begin{center}
        \includegraphics[width=1.0\textwidth]{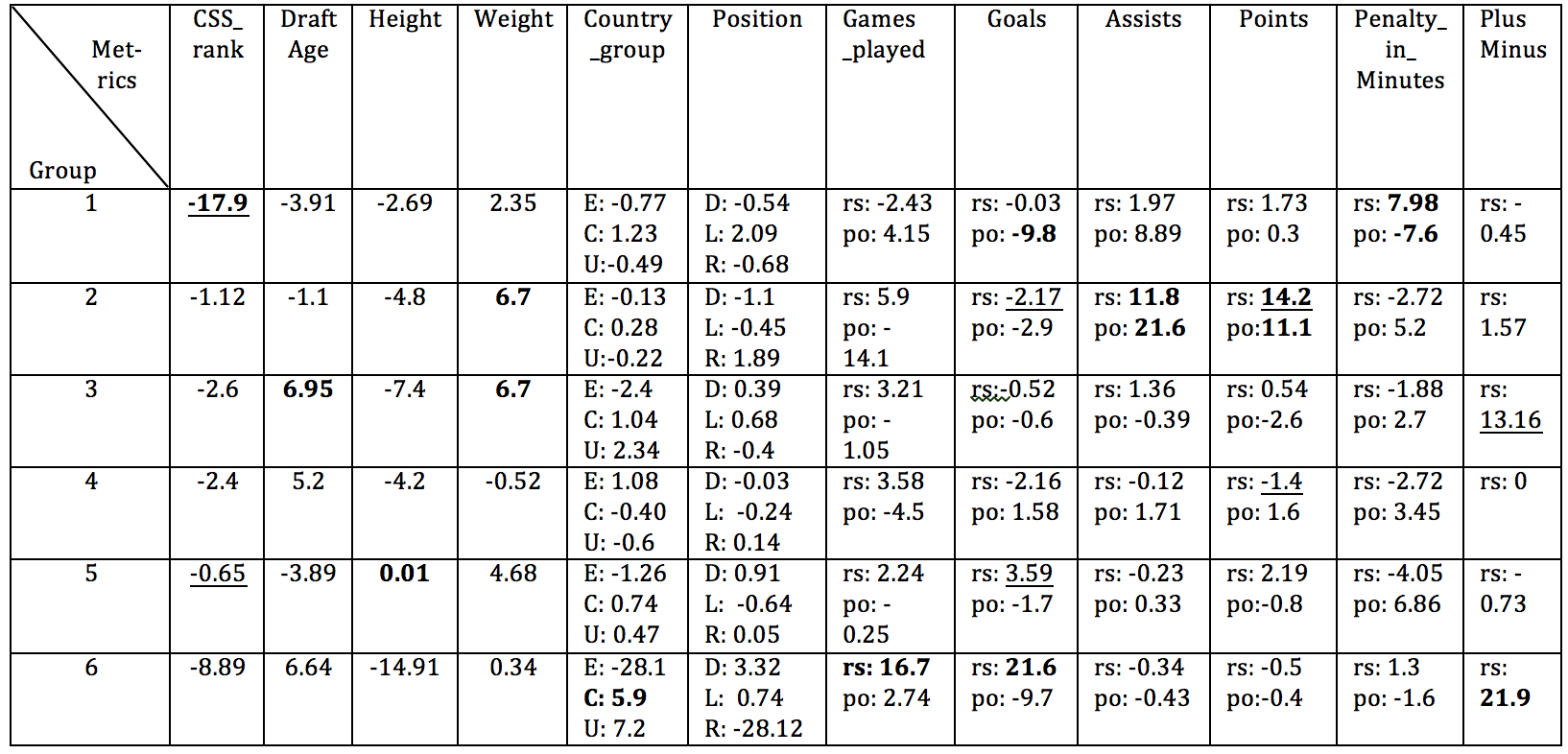}
    \caption{Group 200(4 + 5 + 6 + 7 + 8) Weights Illustration. E = Europe, C = Canada, U = USA, rs = Regular Season, po = Playoff. Largest-magnitude weights are in bold. Underlined weights are discussed in the text.}
    \end{center}
    \end{figure}

    At the tree root, \textit{CSS rank} receives a large negative weight of $-17.9$ for identifying the most successful players in Group 1, where all CSS ranks are better than $12$. Figure 6a shows that the proportion of above-zero to zero-game players decreases quickly in Group 1 with worse CSS rank. However, the decrease is not monotonic. Figure 6b is a scatterplot of the original data for Group 1. We see a strong linear correlation ($p = -0.39$), and also a large variance within each rank. The proportion aggregates the individual data points at a given rank, thereby eliminating the variance. This makes the proportion a smoother dependent variable than the individual counts for a regression model.
    
    \begin{figure}[!h]
    \begin{center}
        \includegraphics[width=1.0\textwidth]{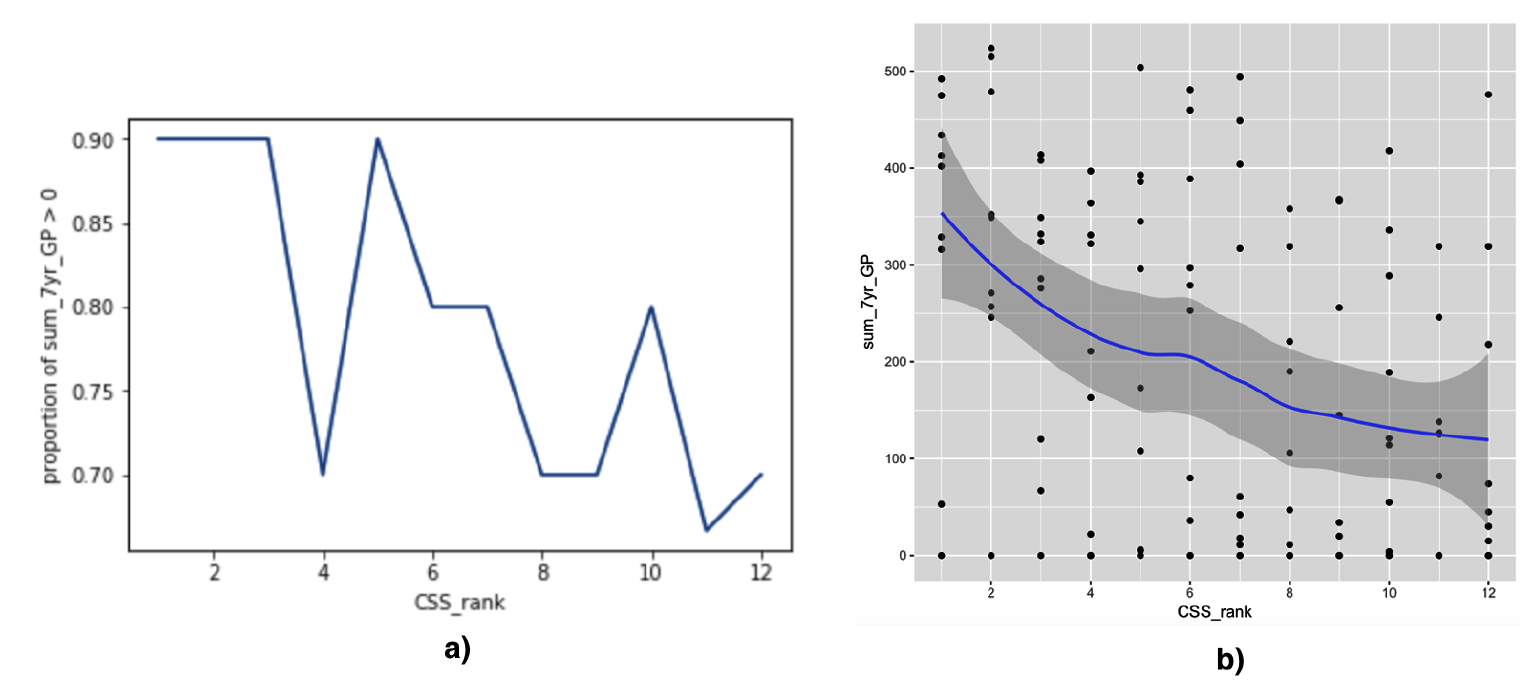}
    \caption{Proportion and scatter plots for CSS\_rank vs. sum\_7yr\_GP in Group 1.}
    \end{center}
    \end{figure}
    
    Group 5 has the smallest logistic regression coefficient of $-0.65$. Group 5 consists of players whose CSS ranks are worse than $12$, regular season points above $12$, and plus-minus above $1$. Figure 7a plots CSS rank vs. above-zero proportion for Group 5. As the proportion plot shows, the low weight is due to the fact that the proportion trends downward only at ranks worse than $200$. The scatterplot in Figure 7b shows a similarly weak linear correlation of $-0.12$. 
    
    \begin{figure}[!h]
    \begin{center}
        \includegraphics[width=1.0\textwidth]{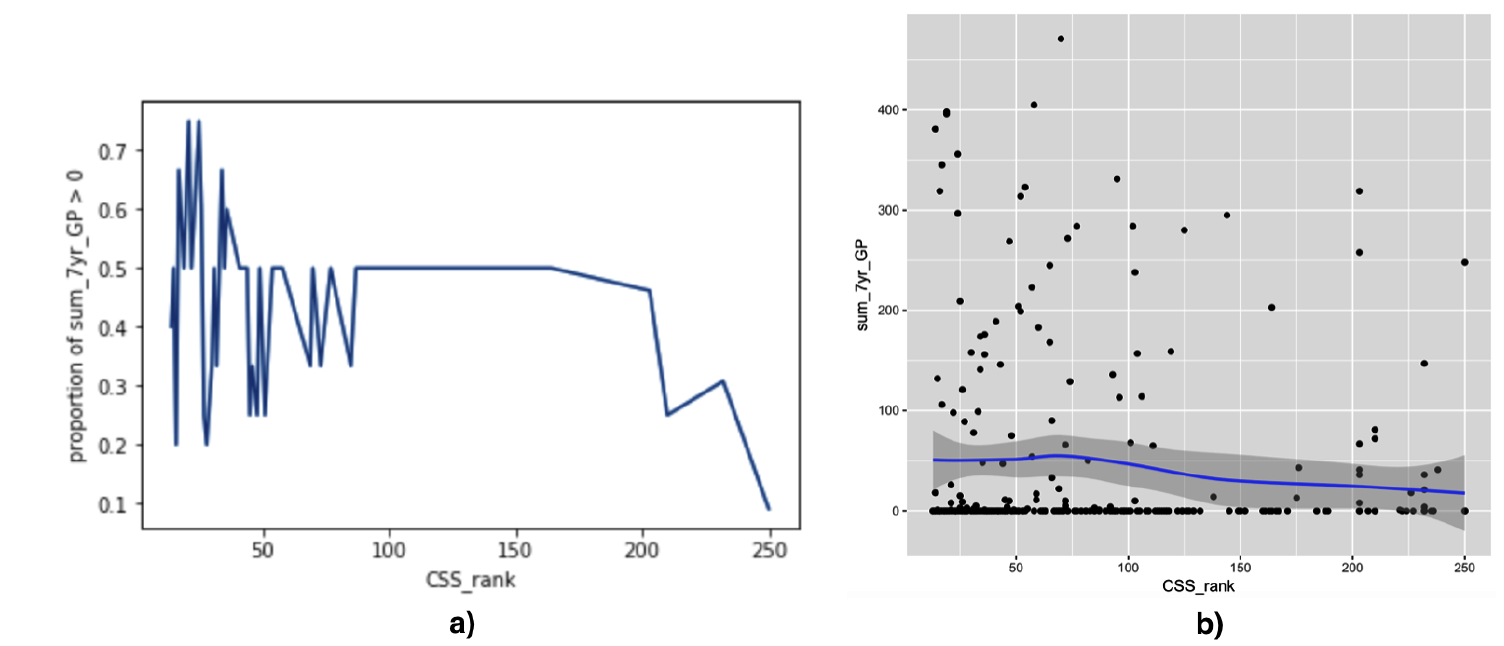}
    \caption{Proportion and scatter plots for CSS\_rank vs.sum\_7yr\_GP in Group 5.}
    \end{center}
    \end{figure}
    
    \textit{Regular season points} are the most important predictor for Group 2, which comprises players with CSS rank worse than $12$, and regular season points below $12$. In the proportion plot Figure 8, we see a strong relationship between points and the chance of playing more than 0 games (logistic regression weight $14.2$). In contrast in Group 4 (overall weight $-1.4$), there is essentially no relationship up to $65$ points; for players with points between $65$ and $85$ in fact the chance of playing more than zero games slightly decreases with increasing points.  
    
    \begin{figure}[!h]
    \begin{center}
        \includegraphics[width=1.0\textwidth]{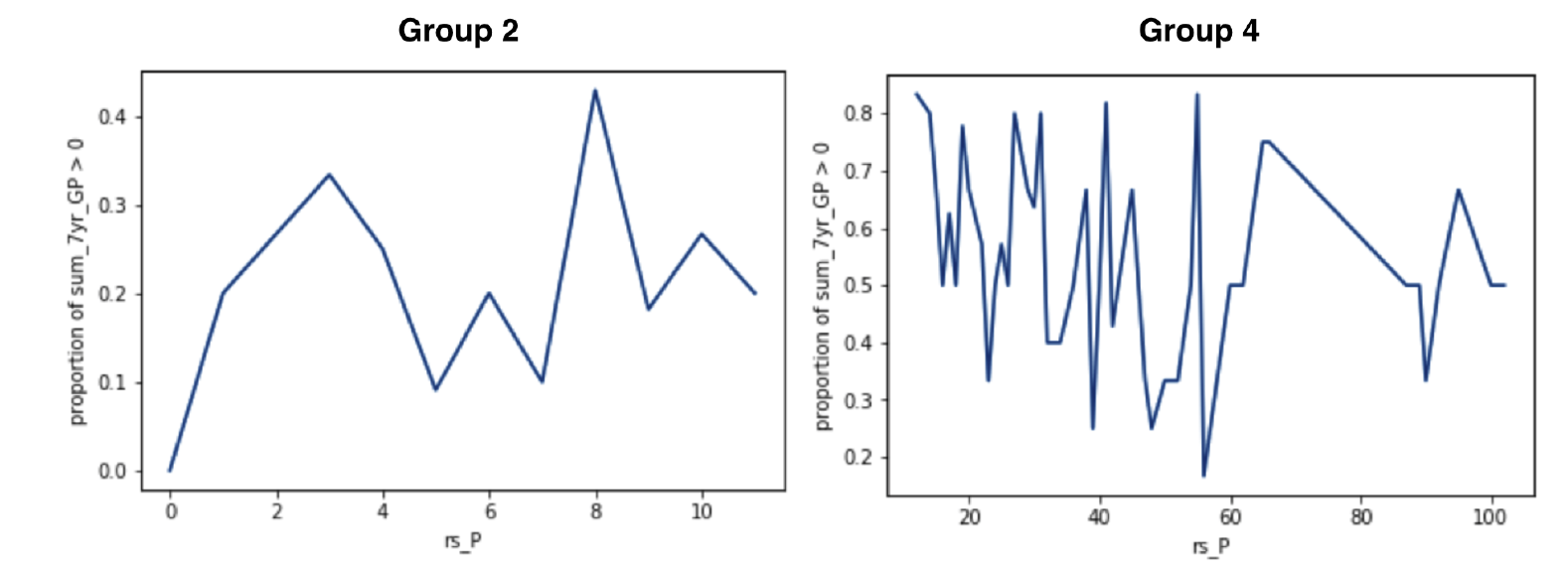}
    \caption{Proportion\_of\_Sum\_7yr\_GP\_greater\_than\_0 vs. rs\_P in Group 2\&4.}
    \end{center}
    \end{figure}
    
    In Group 3, players are ranked at level $12$ or worse, have collected at least $12$ regular season points, and show a negative plus-minus score. The most important feature for Group $3$ is the \textit{regular season plus-minus} score (logistic regression weight $13.16$), which is negative for all players in this group. In this group, the chances of playing an NHL game increase with plus-minus, but not monotonically, as Figure 9 shows. 
    
    \begin{figure}[!h]
    \begin{center}
        \includegraphics[width=1.0\textwidth]{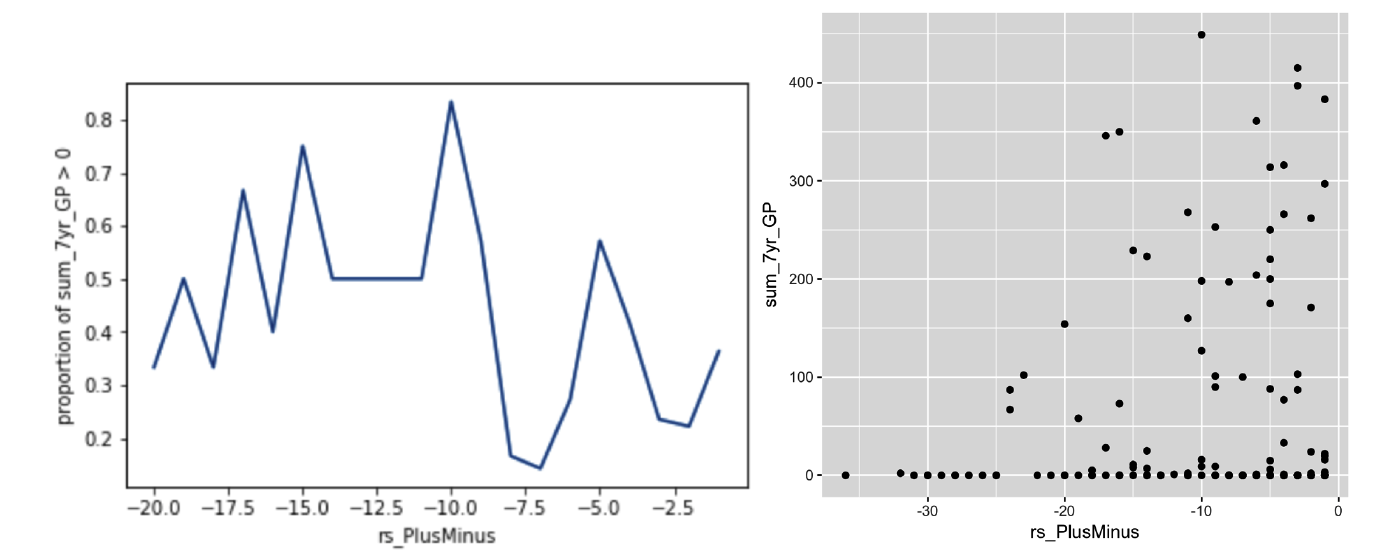}
    \caption{Proportion and scatter plots for rs\_PlusMinus vs.sum\_7yr\_GP in group 3.}
    \end{center}
    \end{figure}
    
    For \textit{regular season goals}, Group 5 assigns a high logistic regression weight of $3.59$. However, Group 2 assigns a surprisingly negative weight of $-2.17$. Group 5 comprises players at CSS rank worse than $12$, regular season points 12 or higher, and positive plus-minus greater than $1$. About $64.8\%$ in this group are offensive players (see Figure 10). The positive weight therefore indicates that successful forwards score many goals, as we would expect. 

    Group 2 contains mainly defensemen ($61.6\%$; see Figure 10). The typical strong defenseman scores $0$ or $1$ goals in this group.  Players with more goals tend to be forwards, who are weaker in this group. In sum, the tree assigns weights to goals that are appropriate for different positions, using statistics that correlate with position (e.g., plus-minus), rather than the position directly.
    
    \begin{figure}[!h]
    \begin{center}
        \includegraphics[width=0.9\textwidth]{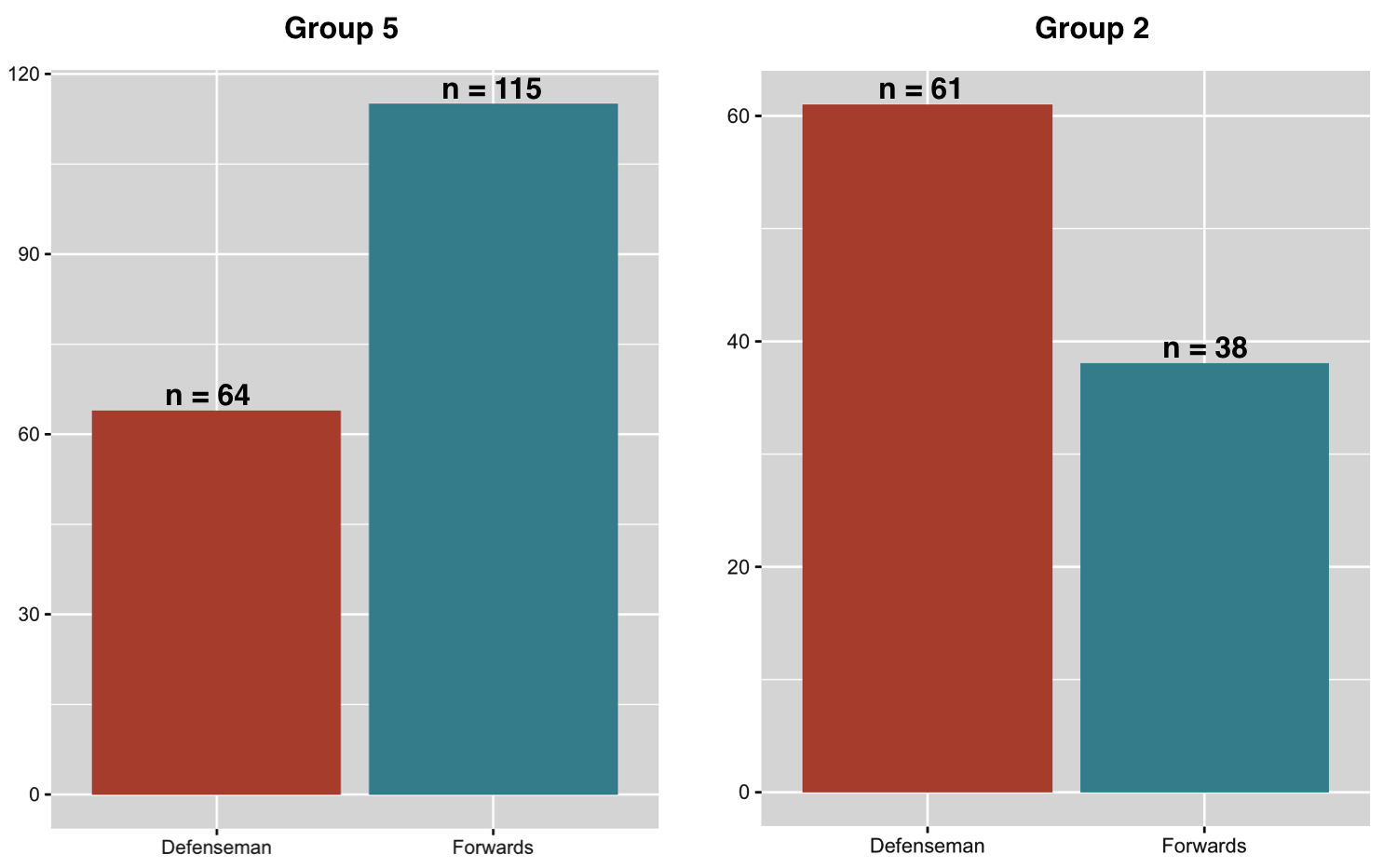}
    \caption{Distribution of Defenseman vs. Forwards in Group 5\&2. The size is denoted as $n$.}
    \end{center}
    \end{figure}

\section{Identifying Exceptional Players}

Teams make drafting decisions not based on player statistics alone, but drawing on all relevant source of information, and with extensive input from scouts and other experts. As Cameron Lawrence from the Florida Panthers put it, \lq the numbers are often just the start of the discussion\rq \cite{Joyce}. In this section we discuss how the model tree can be applied to support the discussion of individual players by highlighting their special strengths. The idea is that the learned weights can be used to identify which features of a highly-ranked player differentiate him the most from others in his group. 

\subsection*{Explaining the Rankings: identify weak points and strong points
} 
Our method is as follows. For each group, we find the average feature vector of the players in the group, which we denote by $\overline{x_{g1}}, \overline{x_{g2}}, ..., \overline{x_{gm}}$ (see Figure 4). We denote the features of player $i$ as $x_{i1}, x_{i2}, ..., x_{im}$ . Then given a weight vector $(w_1, w_m)$ for the logistic regression model of group $g$, the log-odds difference between player $i$ and a random player in the group is given by
    
    \begin{center}
        $\sum_{j=1}^{m}w_j(x_{ij} - \overline{x_{gi}})$
    \end{center}
    
    We can interpret this sum as a measure of how high the model ranks player $i$ compared to other players in his group. This suggests defining as the player's strongest features the $x_{ij}$ that maximize $w_j(x_{ij} - \overline{x_{gi}})$, and as his weakest features those that minimize $w_j(x_{ij} - \overline{x_{gi}})$. This approach highlights features that are $i$) relevant to predicting future success, as measured by the magnitude of $w_j$, and $ii$) different from the average value in the player's group of comparables, as measured by the magnitude of $x_{ij} - \overline{x_{gi}}$. 
    
\subsection*{Case Studies}
Figure 11 shows, for each group, the three strongest points for the most highly ranked players in the group. We see that the ranking for individual players is based on different features, even within the same group. The table also illustrates how the model allows us to identify a group of comparables for a given player. We discuss a few selected players and their strong points. The most interesting cases are often those where are ranking differs from the scouts' CSS rank. We therefore discuss the groups with lower rank first. 

    Among the players who were not ranked by CSS at all, our model ranks \textit{Kyle Cumiskey} at the top. Cumiskey was drafted in place $222$, played $132$ NHL games in his first $7$ years, represented Canada in the World Championship, and won a Stanley Cup in $2015$ with the Blackhawks. His strongest points were being Canadian, and the number of games played (e.g., $27$ playoff games vs. $19$ group average).

    In the lowest CSS-rank group 6 (average $107$), our top-ranked player \textit{Brad Marchand} received CSS rank $80$, even below his Boston Bruin teammate Lucic's. Given his Stanley Cup win and success representing Canada, arguably our model was correct to identify him as a strong NHL prospect.  The model highlights his superior play-off performance, both in terms of games played and points scored. Group 2 (CSS average $94$) is a much weaker group. \textit{Matt Pelech} is ranked at the top by our model because of his unusual weight, which in this group is unusually predictive of NHL participation. In group 4 (CSS average $86$), \textit{Sami Lepisto} was top-ranked, in part because he did not suffer many penalties although he played a high number of games. In group 3 (CSS average $76$), \textit{Brandon McMillan} is ranked relatively high by our model compared to the CSS. This is because in this group, left-wingers and shorter players are more likely to play in the NHL. In our ranking, \textit{Milan Lucic} tops Group 5 (CSS average $71$). At $58$, his CSS rank is above average in this group, but much below the highest CSS rank player (Legein at $13$). The main factors for the tree model are his high weight and number of play-off games played. Given his future success (Stanley Cup, NHL Young Stars Game), arguably our model correctly identified him as a star in an otherwise weaker group. The top players in Group 1 like \textit{Sidney Crosby} and \textit{Patrick Kane} are obvious stars, who have outstanding statistics even relative to other players in this strong group. 

    \begin{figure}[!h]
    \begin{center}
        \includegraphics[width=0.58\textwidth]{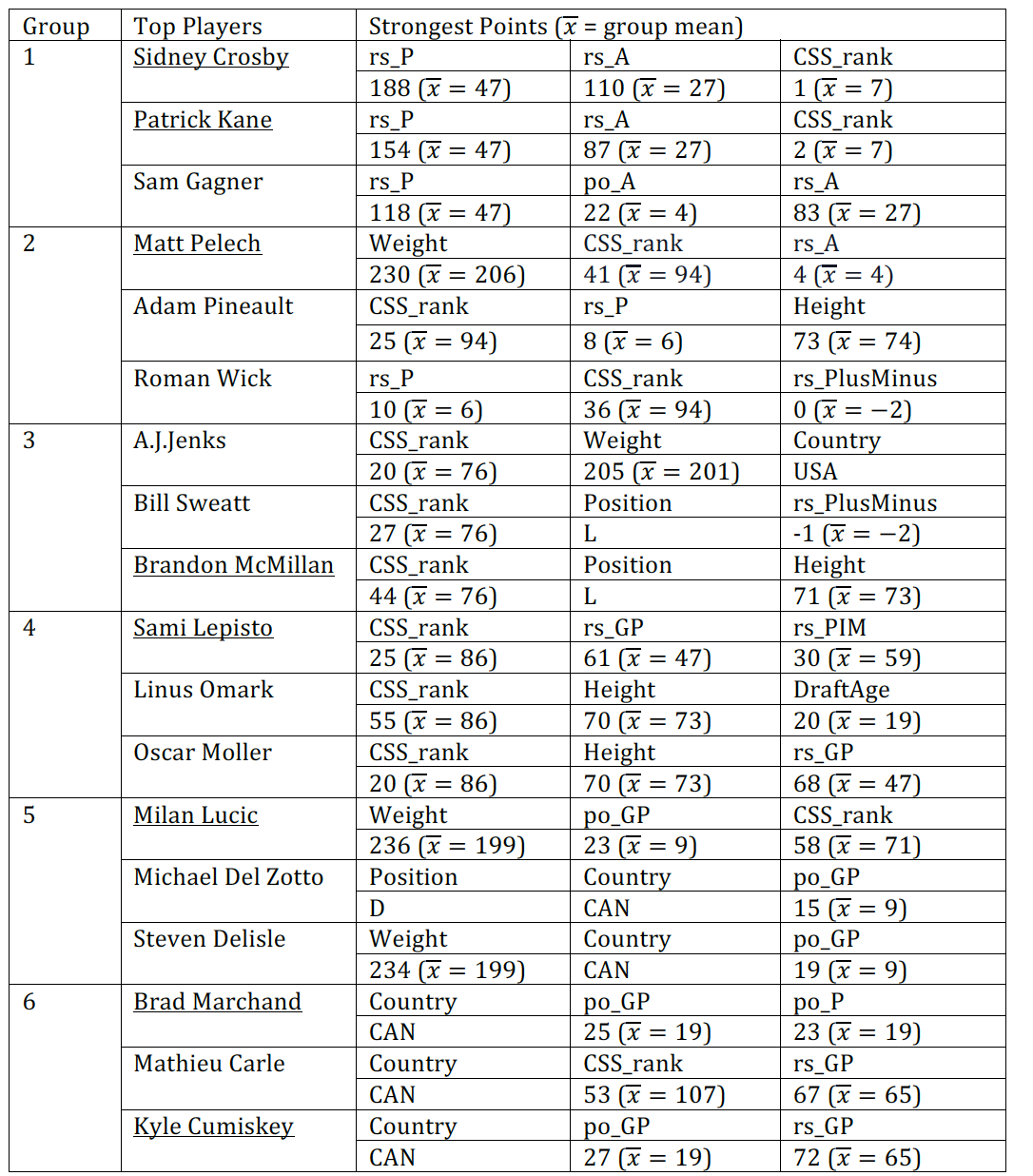}
    \caption{Strongest Statistics for the top players in each group. Underlined players are discussed in the text.}
    \end{center}
    \end{figure}
    
\section{Conclusion and Future Work}
We have proposed building a regression model tree for ranking draftees in the NHL, or other  sports, based on a list of player features and performance statistics. The model tree groups players according to the values of discrete features, or learned thresholds for continuous performance statistics. Each leaf node defines a group of players that is assigned its own regression model. Tree models combine the strength of both regression and cohort-based approaches, where player performance is predicted with reference to comparable players. An obvious approach is to use a linear regression tree for predicting our dependent variable, the number of NHL games played by a player within $7$ NHL years. However, we found that a linear regression tree performs poorly due to the zero-inflation problem (many draft picks never play any NHL game). Instead, we introduced the idea of using a logistic regression tree to predict whether a player plays any NHL game within $7$ years. Players are ranked according to the model tree probability that they play at least $1$ game. 

Key findings include the following. 1) The model tree ranking correlates well with the actual success ranking according to the actual number of games played: better than draft order and competitive with the state-of-the-art generalized additive model \cite{Schuckers2016}. 2) The model predictions complement the Central Scouting Service (CSS) rank. For example, the tree identifies a group whose average CSS rank is only $107$, but whose median number of games played after $7$ years is $128$, including several Stanley Cup winners. 3) The model tree can highlight the exceptionally strong and weak points of draftees that make them stand out compared to the other players in their group.

Tree models are flexible and can be applied to other prediction problems to discover groups of comparable players as well as predictive models. For example, we can predict future NHL success from past NHL success, similar to Wilson \cite{Wilson2016} who used machine learning models to predict whether a player will play more than $160$ games in the NHL after $7$ years. Another direction is to apply the model to other sports, for example drafting for the National Basketball Association.

\bibliographystyle{alpha}
\bibliography{sample}

\begin{appendices} 
	\section*{Spearman Rank Correlation}
Spearman's correlation measures the relevance and direction of monotonic association between two variables \cite{Fieller}. The standard formula for calculating is based on the squared rank differences: 
	
	(1) $p = 1 - \frac{6\sum{d_i^2}}{n(n^2-1)}$, formula for no tied ranks. $n =$ number of ranks, $d_i$ = difference in paired ranks. This is the formula we applied in Table 3.
	
	(2) $p = \frac{\sum_i{(x_i-\overline{x})(y_i-\overline{y})}}{\sqrt{\sum_i(x_i-\overline{x})^2\sum_i(y_i-\overline{y}^2)}}$, where $x_i$ = rank of player $i$ according to ranking $x$, ditto for $y_i$.
	
	Players who have played zero NHL games are tied when ranked by the number of NHL games; this is the only case of ties. Table 3  repeats the calculation of Table 2 using the Pearson correlation among ranks (2) rather than the squared rank differences (1). With this measure also, the model ranking correlates more highly with actual number of games played than the team draft order. 
	
	\begin{table}[!h]
        \centering
        \begin{tabular}{|l|c|c|c|r|}
        \hline
        \begin{tabular}{@{}c@{}} Training Data \\  NHL Draft Years \end{tabular} & \begin{tabular}{@{}c@{}} Out of Sample \\  Draft Years\end{tabular}   & \begin{tabular}{@{}c@{}} Draft Order \\  Pearson Correlation\end{tabular} & \begin{tabular}{@{}c@{}} Tree Model \\  Pearson Correlation\end{tabular} \\ \hline 
        1998, 1999, 2000 & 2001 & 0.43 & 0.69 \\ \hline
        1998, 1999, 2000 & 2002 & 0.45 & 0.72 \\ \hline
        2004, 2005, 2006 & 2007 & 0.48 & 0.60 \\ \hline
        2004, 2005, 2006 & 2008 & 0.51 & 0.58 \\ \hline

        \end{tabular}
        \caption{Pearson Correlation of NHL ranks.}
       
    \end{table}

\section*{LogitBoost Algorithm}
Ensemble methods provide a combination of classifiers to obtain a better predictive result than any of the standalone constituents \cite{Ryder,Freund,Cherkauer}. While a model tree constructs a set of different classifiers, it also partitions the space of players, so the prediction for each player is based on exactly one model only, rather than a weighted majority vote. The LogitBoost algorithm \cite{Friedman00} combines a model tree with weighted votes by building a separate model for each tree node (including non-leaf nodes). The prediction for a specific player is then the weighted vote of all models along the branch assigned to the player. This offers some of the advantages of a hierarchical shrinkage model in smoothing parameter values so that predictions from players from similar groups tend to use similar weights. In this paper, we used the tree structure learned from LogitBoost, with simple maximum likelihood estimates for the weights to make the weight for a leaf more interpretable by fitting the data in the leaf’s group more closely. The GUIDE system \cite{GUIDE} is a well-developed software package that supports building ensembles of model trees. While such an ensemble tends to have even higher predictive accuracy, we have in this paper built only a single model tree to maintain interpretability.

\end{appendices}

\end{document}